\documentclass[11pt]{article}

\usepackage[preprint]{acl}

\usepackage{times}
\usepackage{latexsym}

\usepackage[T1]{fontenc}

\usepackage[utf8]{inputenc}

\usepackage{microtype}

\usepackage{inconsolata}

\usepackage{graphicx}

\usepackage{float}

\usepackage{graphicx}
\usepackage{amsmath}
\usepackage{amssymb}
\usepackage{mathtools}
\usepackage{amsthm}

\usepackage{tcolorbox}
\tcbuselibrary{skins}
\usepackage{lipsum}
\usepackage{algorithm}
\usepackage{algpseudocode}
\usepackage{marvosym}
\usepackage{booktabs}

\usepackage{amsmath,amssymb,amsthm}
\usepackage{listings}
\usepackage{enumitem}

\theoremstyle{plain}

\theoremstyle{definition}

\theoremstyle{remark}

\title{Fuzzy Categorical Planning: Autonomous Goal Satisfaction with Graded Semantic Constraints}

\author{Shuhui Qu \\
  Stanford University \\
  \texttt{shuhuiq@stanford.edu} \\
  }

\begin{document}
\maketitle

\begin{abstract}
Natural-language planning often involves \emph{vague} predicates (e.g., \textit{suitable substitute}, \textit{stable enough}) whose satisfaction is inherently graded. Existing category-theoretic planners provide compositional structure and pullback-based hard-constraint verification, but treat applicability as crisp, forcing thresholding that collapses meaningful distinctions and cannot track quality degradation across multi-step plans.
We propose \emph{Fuzzy Category-theoretic Planning} (FCP), which annotates each action (morphism) with a degree in $[0,1]$, composes plan quality via a t-norm (Łukasiewicz), and retains crisp executability checks via pullback verification. FCP grounds graded applicability from language using an LLM with $k$-sample median aggregation and supports meeting-in-the-middle search using residuum-based backward \emph{requirements}.
We evaluate on (i) public PDDL3 preference/oversubscription benchmarks and (ii) \textsc{RecipeNLG-Subs}, a missing-substitute recipe-planning benchmark built from RecipeNLG with substitution candidates from Recipe1MSubs and FoodKG. FCP improves success and reduces hard-constraint violations on \textsc{RecipeNLG-Subs} compared to LLM-only and ReAct-style baselines, while remaining competitive with classical PDDL3 planners. 
\end{abstract}

\section{Introduction}
Natural-language instructions for everyday tasks are often \emph{vague} rather than strictly logical: \emph{use a reasonable substitute}, \emph{make it stable enough}, \emph{finish soon}. Such predicates are not merely noisy measurements of a hidden binary condition; they are frequently \emph{graded} in meaning, admitting degrees of satisfaction \citep{zadeh1965fuzzy,williamson2002vagueness,kennedy2007vagueness}. In practice, however, most planning formalisms operation applicability as binary \citep{ghallab2004automated,aeronautiques1998pddl,fox2003pddl2}. This forces planners to either (i) reject near-feasible actions whose satisfaction falls just below a cutoff, or (ii) threshold and treat all actions above the cutoff as equivalent, collapsing distinctions between marginal and clearly good options \citep{bistarelli1999semiring,smith2004choosing,qu2025category}.

A natural question is whether a large language model (LLM) already ``handles'' vagueness implicitly \citep{brown2020language,wei2022chain}. LLMs are often effective at \emph{local} judgments (e.g., whether an ingredient is a plausible substitute) \citep{wiggins2022opportunities,ouyang2022training}, but planning requires \emph{compositional} use of such judgments across multi-step sequences under hard constraints \citep{valmeekam2023planbench,valmeekam2023planning,huang2022language}. Purely neural planning typically provides no explicit, controllable account of how partial satisfaction compounds across steps, and its implicit thresholds can be difficult to reproduce or audit across prompts and model versions \citep{li2023theory,li2023explanation,arora2022ask}. Our aim is therefore modest: we use the LLM only as a grounding module for graded predicates, while keeping executability checks and plan-level acceptance explicit and symbolic \citep{qu2025category,ghallab2004automated}.

This paper asks: \textbf{can we extend compositional planning to reason over graded satisfaction while preserving crisp executability?}
We build on categorical planning, where actions are morphisms, plans are compositions, and pullback-style checks enforce hard constraints in a verifiable way \citep{qu2025category,spivak2013category,rydeheard1988computational}. Existing categorical planners are crisp: a morphism either applies or it does not \citep{qu2025category}. This is well-matched to hard resources and logical preconditions \citep{ghallab2004automated,fox2003pddl2}, but it is mismatched to vagueness, where the intended semantics are inherently graded \citep{zadeh1965fuzzy,williamson2002vagueness,kennedy2007vagueness}.

\paragraph{Approach.}
We propose \emph{Fuzzy Category-theoretic Planning} (FCP), a fuzzy enrichment of categorical planning. Each action (morphism) is annotated with an applicability degree in $[0,1]$ capturing how well it satisfies vague requirements in the current state. These degrees are grounded from natural language using an LLM with $k$-sample aggregation, while resources, logic, and temporal constraints remain crisp and are enforced via the same pullback-style verification as prior work. Plan quality is computed compositionally via a t-norm (Łukasiewicz in our main instantiation), yielding an explicit and interpretable mechanism for tracking quality degradation across steps. FCP returns an executable plan only after explicit $\alpha$-cut verification of both (i) crisp feasibility and (ii) required plan quality.

\paragraph{Evaluation.}
We evaluate FCP in two settings where fuzziness is \emph{planning-relevant} rather than a one-shot classification problem.
First, we use public PDDL3 benchmarks with preferences and oversubscription-style objectives, which provide a standard mechanism for soft desiderata while preserving hard executability constraints \cite{gerevini2009deterministic,menkes2004effective}. Second, we construct \textsc{RecipeNLG-Subs} by converting RecipeNLG recipes into missing-ingredient planning tasks that require selecting substitutes and updating procedures accordingly \cite{bien2020recipenlg}. Substitution candidates and constraint propagation are derived from a food knowledge graph and recipe-oriented substitution resources \cite{haussmann2019foodkg,fatemi2023learning}. We evaluate by (i) hard-constraint violation / success using validators and checkers, and (ii) BLEU against canonicalized references for recipe realizations.

\paragraph{Contributions.}
(1) We introduce a fuzzy-enriched categorical planning formulation that propagates graded satisfaction compositionally while preserving crisp executability checks.
(2) We integrate LLM-based grounding with symbolic pullback verification and explicit $\alpha$-cut plan acceptance, making plan quality controllable and auditable.
(3) We evaluate on PDDL3 preference/oversubscription planning and a new RecipeNLG-based missing-substitute benchmark, isolating the role of fuzzy composition under hard constraints.
(4) We provide mechanism tests (aggregation, t-norm choice, and acceptance policy) to clarify which components are most responsible for improvements.

The rest of the paper is organized as follows.
Section~\ref{sec:fcp} introduces Fuzzy Category-theoretic Planning (FCP), including fuzzy enrichment, LLM-based grounding, bidirectional search, and pullback-style crisp verification.
Section~\ref{sec:experiments} describes the benchmarks (PDDL3 preferences/oversubscription and RecipeNLG-Subs), baselines, and evaluation protocol.
Section~\ref{sec:results} presents the main results, mechanism tests, degradation analyses, and backbone comparisons.
\section{Related Work}

\subsection{Classical Planning, Preferences, and Oversubscription}

Classical symbolic planners provide strong correctness guarantees by searching over formally specified transition systems \citep{ghallab2004automated,helmert2006fast,richter2010lama}. To model \emph{non-mandatory} desiderata while preserving hard executability, the planning community has developed preference and soft-constraint formalisms, notably PDDL3 preferences \citep{gerevini2009deterministic}. A closely related line is \emph{oversubscription planning}, where the agent maximizes achieved utility when not all goals can be simultaneously satisfied \citep{smith2004choosing}; several works show how soft goals can be compiled into standard cost-based planning (under suitable assumptions), enabling reuse of classical planners \citep{keyder2009soft}. Recent formulations further connect oversubscription to multi-cost classical planning, clarifying algorithmic trade-offs and benchmarks \citep{katz2019oversubscription}.

These approaches treat softness as \emph{preferences over crisp predicates} (e.g., which goals to achieve), whereas our focus is \emph{graded satisfaction of a predicate itself} (e.g., how well an action meets “suitable substitute” in context). In our setting, the plan remains required to be \emph{hard-feasible}, but its quality is computed compositionally from graded applicability.

\subsection{Neural and LLM-Based Planning}

Large language models (LLMs) enable planning from natural-language instructions without fully specified symbolic domain models, and have been used for zero-/few-shot action planning and instruction following \citep{huang2022language,ahn2022can}. However, multiple studies show that LLM-generated plans can fail under compositional constraints and long-horizon dependencies, often producing invalid or non-executable sequences even when individual steps look plausible \citep{valmeekam2023planbench, valmeekam2023planning}. Recent work also explores translating natural language into planning formalisms (e.g., PDDL-style representations) to regain executability, but this typically restores \emph{binary} constraint semantics once compiled \citep{gestrin2024nl2plan}.

Our framework takes a complementary stance: we use the LLM only as a \emph{grounding module} for vague predicates (to estimate graded applicability), while keeping feasibility checks and plan acceptance explicit and auditable.

\subsection{Search-Augmented and Structured Reasoning for Planning}

To mitigate brittleness of direct generation, several methods add structured exploration. Tree-structured deliberation frameworks such as Tree-of-Thoughts \citep{yao2023tree} and tool-augmented decomposition schemes such as ReWOO \citep{xu2023rewoo} improve reliability by expanding and selecting among candidate reasoning branches, but they still rely on largely \emph{binary} accept/reject signals at each branch. Planning-specific tree search has also been proposed, including MCTS-style formulations that use learned (or LLM-derived) heuristics to guide expansion \citep{hao2023reasoning}. Separately, bidirectional search provides a well-known complexity reduction from $O(b^L)$ to $O(b^{L/2})$ in many planning settings; recent work has revived goal-directed bidirectional search in domains such as retrosynthesis and synthesis planning \citep{lan2025retro,yu2024double}.

FCP is compatible with these developments but differs in what is propagated: we propagate \emph{graded satisfaction} through composition (forward) and propagate \emph{requirements} via residuation (backward), so that meeting-in-the-middle is governed by an explicit notion of plan-quality feasibility rather than an implicit threshold.

\subsection{Fuzzy Logic and Graded Satisfaction in Planning}

Fuzzy logic formalizes graded membership and provides algebraic operators for combining partial satisfaction \citep{zadeh1965fuzzy}. Early work in robotics and control used fuzzy rule bases and fuzzy behaviors to act under imprecise concepts and sensor uncertainty \citep{saffiotti1997uses}. In mainstream planning languages, continuous fluents are supported (e.g., PDDL2.1) \citep{fox2003pddl2}, and richer hybrid dynamics are captured in extensions such as PDDL+ \citep{fox2006exploration}; however, numeric fluents are typically treated as quantities to optimize or constrain, not as \emph{graded truth values} for vague predicates. Classical fuzzy dynamic programming and related fuzzy optimal control frameworks optimize over graded notions of value, but common multiplicative aggregations can make long-horizon composition difficult to interpret \citep{yao1998comparative}.

Our work revisits fuzzy planning from a different angle: we aim to represent graded applicability while retaining \emph{hard} executability and compositional verification. Concretely, we integrate fuzzy enrichment into a category-theoretic planning backbone, and use \L{}ukasiewicz composition to make degradation explicit and easy to audit.

\subsection{Soft Constraints vs.\ Graded Predicates}

Soft-constraint and semiring-based formalisms generalize constraint satisfaction and support preferences through algebraic aggregation \citep{bistarelli1999semiring,wilson2004extending}. Preference-based planning and oversubscription planning optimize which goals or preferences to satisfy under hard feasibility \citep{gerevini2009deterministic,smith2004choosing}.

The key distinction is representational: these methods model \emph{softness over crisp propositions} (“achieve goal $g$ if possible”), whereas we model \emph{graded satisfaction of the predicate itself} (“action $f$ satisfies vague precondition $p$ to degree $\mu$”). This matters for language planning because many instruction predicates (e.g., “reasonable substitute”, “stable enough”) are naturally graded rather than merely preferred.

\subsection{Category-Theoretic Structure for Planning and Reasoning}

Category theory provides a compositional language for systems and transformations, and has been applied broadly in computer science and AI as a unifying semantics for structured reasoning \citep{rydeheard1988computational,spivak2013category}. Qu et al.\ introduced category-theoretic planning where actions are morphisms, plans are compositions, and pullback-style checks enforce compatibility of resources and constraints while supporting efficient search \citep{qu2025category}. Their framework, however, assumes crisp applicability (a morphism applies or not), which is well-suited to hard constraints but less suited to vague predicates.

We build on this foundation by enriching morphisms with graded applicability while keeping pullback verification for hard constraints unchanged. This yields a planner that can represent and propagate semantic vagueness without giving up executability guarantees.

\subsection{Vagueness in Natural Language and LLM Uncertainty}

Linguistics distinguishes semantic vagueness (gradable predicates such as “tall”) from epistemic uncertainty about hidden boundaries \citep{williamson2002vagueness,kennedy2007vagueness}. In contrast, most LLM uncertainty work targets calibration for factual correctness or epistemic confidence (e.g., “does the model know that it knows?”), which is conceptually different from graded predicate meaning \citep{kadavath2022language,kuhn2023semantic}. Our goal is to adapt semantic vagueness in planning via explicit graded composition and controllable acceptance thresholds, rather than to estimate epistemic confidence.

\section{Fuzzy Category-Theoretic Planning}
\label{sec:fcp}

We extend categorical planning to reason compositionally over graded predicate satisfaction. Natural-language instructions often rely on vague concepts such as \textit{suitable substitute} or \textit{stable enough} that resist binary classification. Our goal is modest: we aim to make these graded judgments \emph{explicit}, \emph{composable}, and \emph{auditable}, while keeping hard feasibility constraints crisp and verifiable.

\subsection{Fuzzy-Enriched Planning Category}

\paragraph{State representation.}
We represent a state as $w=(r,s,l,t)$, where $r$ denotes resources, $s$ a symbolic description of progress, $l$ a set of logical constraints, and $t$ temporal allocations. Throughout, we treat $(r,l,t)$ as \emph{hard} (crisp) constraints: any violation in these components renders a transition infeasible, independent of graded scores.

\paragraph{Fuzzy morphisms.}
An action is a morphism $f:w\to w'$ equipped with an applicability degree $\mu(f;w)\in[0,1]$ that measures how well $f$ satisfies \emph{graded} preconditions in context $w$. For preconditions that are intrinsically crisp, we simply use $\mu(f;w)\in\{0,1\}$. In our implementation, a candidate action is accepted as a feasible successor only if it passes crisp checks on $(r,l,t)$; fuzzy applicability does not override hard feasibility.

\paragraph{Łukasiewicz composition.}
We compose graded applicability using the t-norm and its residuum:
\begin{equation}
    \begin{split}
        a\otimes b \;&=\; \max(0,a+b-1), \\
        a \Rightarrow b \;&=\; \min(1,1-a+b).
    \end{split} 
\end{equation}

Given a plan $\pi=(f_1,\ldots,f_n)$ producing states $w_0,\ldots,w_n$, we propagate state membership by
\begin{equation}
    \begin{split}
        \mu(w_i)&=\mu(w_{i-1})\otimes \mu(f_i;w_{i-1}),\\ &where, \mu(w_0)=1
    \end{split}
\end{equation}

and define the plan membership as $\mu(\pi)=\mu(w_n)$. For Łukasiewicz, the iterated t-norm admits the closed form
\begin{equation}
    \begin{split}
        \mu(\pi)
        \;&=\;
        \bigotimes_{i=1}^n \mu(f_i;w_{i-1})\\
        \;&=\;
        \max\!\Bigl(0,\sum_{i=1}^n \mu(f_i;w_{i-1})-(n-1)\Bigr).
    \end{split}
\end{equation}

This expression makes degradation explicit: for example, six actions with $\mu_i=0.8$ yield $\mu(\pi)=\max(0,4.8-5)=0$.

\paragraph{Why Łukasiewicz?}
We adopt Łukasiewicz primarily because it provides (i) a nilpotent, closed-form degradation law (above), (ii) a strong residuum $a\Rightarrow b$ that supports backward ``requirement'' propagation, and (iii) an interpretable, linear accumulation of degradation (in contrast to exponential decay under product t-norm or no decay under Gödel/min). We also compare alternative t-norms in ablation.

\subsection{Grounding Vague Predicates}
\label{sec:grounding}

Vague predicates such as \textit{suitable} or \textit{enough} require a numerical membership in $[0,1]$ for each candidate action in context. We use a language model as a \emph{membership oracle} that produces a graded judgment conditioned on the current state description and the predicate rubric. Concretely, for a morphism $f:w\to w'$ and a vague predicate $p$, we draw $k$ independent samples
\[
\hat{\mu}_1,\ldots,\hat{\mu}_k \in [0,1]
\]
by querying the same prompt template under stochastic decoding (temperature $T>0$), and then aggregate them into a single membership score
\[
\mu(f;w) \;=\; \mathrm{Agg}_k(\hat{\mu}_{1:k}).
\]
We refer to $k$ as the \emph{aggregation width}.

\paragraph{Aggregation rule.}
Unless stated otherwise, we set $\mathrm{Agg}_k$ to the median:
\[
\mu(f;w) \;=\; \mathrm{median}\bigl(\hat{\mu}_1,\ldots,\hat{\mu}_k\bigr),
\]
using an odd $k$ (default $k=5$). The median is robust to occasional outlier responses and empirically yields more stable memberships than a single query. We treat $k$ as a tunable parameter controlling a simple cost--stability trade-off; in the experiments we report results for the default $k$ and include an ablation over $k$.

\paragraph{Prompting.}
Each query asks the model to rate how well the current state supports the predicate for executing action $f$, using a calibrated 0--100 scale with semantic anchors (0=completely unsuitable, 50=marginal, 100=perfect) and domain-specific factors (e.g., cooking: flavor compatibility, texture, dietary constraints; assembly: stability, clearance, material strength). We parse the returned scalar and normalize to $[0,1]$.

\paragraph{Scope.}
We emphasize that the LLM is used only to produce local membership judgments; hard constraints and final plan acceptance remain explicit and symbolic (via pullback-based verification and $\alpha$-cut checking). Our claims about performance therefore concern the effect of \emph{aggregated} memberships on downstream fuzzy composition rather than assuming the LLM alone resolves vagueness reliably in multi-step planning.




\subsection{Bidirectional Search with Fuzzy Propagation}

We adapt bidirectional AND-OR search \cite{yu2024double} to propagate fuzzy memberships while retaining the usual $O(b^{L/2})$ complexity benefit in the branching factor $b$ and plan length $L$.

\paragraph{Forward propagation.}
From an initial state $w_0$ with $\mu(w_0)=1$, we expand a forward frontier. For each expanded state $w$ with membership $\mu(w)$, we propose candidate actions $f:w\to w'$, obtain $\mu(f;w)$ via grounding, and apply crisp feasibility checks to ensure $w'$ satisfies all hard constraints in $(r,l,t)$. If feasible, we propagate:
\[
\mu(w')=\mu(w)\otimes \mu(f;w).
\]

\paragraph{Backward propagation (requirements).}
To support meeting-in-the-middle, we maintain a \emph{requirement function} $r_B(\cdot)$ that lower-bounds the forward membership needed at a state in order to still achieve a target goal quality. We initialize $r_B(w^*)=\alpha$ for goal $w^*$ and define:
\[
r_B(w)=\max_{f:w\to w'} \bigl(\mu(f;w)\Rightarrow r_B(w')\bigr).
\]
The $\max$ corresponds to the existential nature of planning: it suffices that there exists an action leading to a feasible continuation. The residuum is used as an algebraic inversion of Łukasiewicz composition: from $a\otimes x\ge b$ we can derive the sufficient condition $x\ge a\Rightarrow b$.

\paragraph{Search guidance.}
We guide frontier expansion using a learned distance $D:W\times W\to[0,1]$ combining symbolic, resource, logical, and temporal components:

\begin{equation}
    \begin{split}
        D(w_1,w_2)=&\alpha_s d_s(s_1,s_2)+\alpha_r \|r_1-r_2\|\\
        &+\alpha_l d_l(l_1,l_2)+\alpha_t d_t(t_1,t_2),
    \end{split}
\end{equation}

where $d_s$ is learned by supervised regression on state pairs extracted from completed search trees, and the remaining terms are simple dissimilarities. We emphasize that $D$ is used only as a heuristic for prioritization; it does not enter correctness or acceptance checks.

\paragraph{Termination.}
We propose meet candidates $(w_F,w_B)$ when $D(w_F,w_B)\le \varepsilon_D$. A candidate is accepted only if it also satisfies (i) crisp pullback compatibility and (ii) the forward membership meets the backward requirement:
\begin{equation}
    \begin{split}
        D(w_F,w_B)\le \varepsilon_D,\\
\textsc{PullbackCompatible}(w_F,w_B),\\
\mu(w_F)\ge r_B(w_F).
    \end{split}
\end{equation}

\subsection{Pullback Verification}

Given a candidate meet $(w_F,w_B)$, we verify compositional validity via the same pullback-style crisp compatibility checks as \citet{qu2025category}. Concretely, we check that there exists a merged state $w_p$ whose resource, logical, and temporal components jointly satisfy all hard constraints induced by both partial plans. If any crisp constraint is violated (resource infeasibility, logical inconsistency, temporal conflict), the meet is rejected regardless of fuzzy membership. When the crisp merge succeeds, we assign the merged state's membership by the same Łukasiewicz composition used elsewhere.

\subsection{Plan Acceptance via $\alpha$-Cuts}

Execution ultimately requires a discrete plan. We accept a plan $\pi$ only if:
\[
\mu(\pi)\ge \alpha
\quad\text{and all crisp constraints hold.}
\]
We view $\alpha$ as an explicit, user- or domain-chosen quality threshold. In some settings, it is helpful to adapt $\alpha$ by context and plan length:
\[
\alpha=\alpha_{\text{base}}\cdot f_{\text{criticality}}\cdot f_{\text{length}},
\]
where $\alpha_{\text{base}}$ is a small set of interpretable presets (e.g., casual/typical/important/critical) and $f_{\text{length}}$ reduces the threshold for longer plans to reflect inevitable degradation under nilpotent composition. In our experiments, we report both a fixed-$\alpha$ variant and an adaptive-$\alpha$ variant and treat this policy as an ablatable component.

\subsection{Formal Properties}

We summarize the properties as follows:

\paragraph{Soundness (with respect to hard constraints and thresholding).}
If the method returns a plan $\pi$, then (i) $\pi$ satisfies all hard constraints because every transition and merge is validated by crisp feasibility and pullback compatibility checks, and (ii) $\mu(\pi)\ge \alpha$ by explicit evaluation. We do not claim optimality under any fuzzy ordering.

\paragraph{Complexity.}
Fuzzy operations (t-norm composition, residuum, membership lookup) are $O(1)$ per action application and do not change the asymptotic $O(b^{L/2})$ bidirectional search complexity.

\paragraph{Degradation law}
For Łukasiewicz composition, plan membership is
\[
\mu(\pi)=\max\!\Bigl(0,\sum_i \mu_i-(n-1)\Bigr),
\]
so quality degradation is intrinsic to the structure rather than an artifact of the search procedure.

\subsection{Addressing Nilpotent Degradation}
\label{sec:mitigation}

Nilpotency places practical limits on plan depth when step-wise memberships are imperfect. We use three pragmatic mitigations that remain within the same framework:

\paragraph{Semantic chunking.}
We optionally replace short, semantically coherent action sequences with macro-morphisms, reducing effective plan length. Macro-morphisms are assigned memberships either by a dedicated grounding query over the chunk or by empirical success estimates when available.

\paragraph{Calibration from execution.}
When execution outcomes can be observed, we adjust LLM-elicited memberships toward observed success rates using a bounded update rule. This aims to correct systematic biases (e.g., overly conservative grounding) without changing the underlying t-norm semantics.

Together, these mitigations broaden the regime where fuzzy enrichment is useful, while keeping hard feasibility checks unchanged.

\section{Experiments}
\label{sec:experiments}
We evaluate \textsc{Fuzzy Category-theoretic Planning} (FCP) with the goal of understanding when graded composition is helpful for \emph{planning} problems where (i) feasibility remains crisp but (ii) objectives are non-binary. Concretely, we focus on two types of vagueness: \textbf{(A)} soft goals / preferences and, and \textbf{(B)} missing-substitute recipe planning.

\subsection{Benchmarks}
\label{sec:benchmarks}

\paragraph{(A) PDDL3 Preferences \& Oversubscription Suite.}
We evaluate on publicly available PDDL3 benchmark problems that include \texttt{preferences} and a plan metric trading off hard costs with preference violations (e.g., terms involving \texttt{(is-violated \textit{name})}). In PDDL3, preferences are \emph{soft constraints}: they do not change goal truth, but contribute to the plan metric through violation counts. This makes PDDL3 a standard way for oversubscription planning where the planner must remain feasible while optimizing soft desiderata. 

\paragraph{(B) \textsc{RecipeNLG-Subs}: RecipeNLG with Missing-Substitute Planning Actions.}
We construct a recipe-planning benchmark by converting RecipeNLG recipes (ingredient list + procedural steps) into \emph{missing-substitute} tasks: each task removes one or more key ingredients and requires the planner to (i) select substitutes and (ii) update the ingredient list and procedural steps accordingly. Substitution candidates are obtained from Recipe1MSubs; ingredient canonicalization and constraint propagation (e.g., dietary tags, ingredient categories, incompatibilities) are supported by FoodKG.
We report dataset construction details and filtering in Appendix.

\paragraph{Instance controls.}
To make fuzziness \emph{planning-relevant} rather than a one-shot classification problem, we stratify by:
(i) number of missing ingredients ($m$),
(ii) ambiguity level (number of plausible substitutes),
(iii) plan length (number of steps).

\subsection{Methods Compared}
\label{sec:baselines}
We compare FCP against (i) \emph{classical planners} that are strong in crisp validity and metric optimization, and (ii) \emph{LLM-based} baselines that excel in fluent editing but may violate constraints.

We include PDDL3-capable planners commonly used in preference-based / metric evaluations \citep{gerevini2009deterministic}:

\begin{itemize}
    \item \textbf{SGPlan5}:  a heuristic planner with strong empirical performance on IPC-styles tasks.
    \item \textbf{MIPS-XXL}: metric-focused planners supporting preference domains.
    \item \textbf{LPPG)}:  additional reference systems used in IPC5 preference/metric evaluations.
\end{itemize}
These systems provide a strong reference for (i) near-zero hard-constraint violation and (ii) optimization of preference satisfaction.

We also include LLM-based baselines (Default GPT-4o):
\begin{itemize}
    \item \textbf{LLM (Direct Prompt)}: generate the full recipe from LLM model directly.
    \item \textbf{SubDB-Retrieve (RAG)}: retrieve top substitute(s) from Recipe1MSubs conditioned on ingredient type and constraints, then apply template edits to steps.
    \item \textbf{ReAct (tool-augmented)}: interleave reasoning and tool calls.
\end{itemize}

\paragraph{Our method: FCP (fuzzy categorical planning).}
FCP uses (i) LLM-based membership grounding with $k$-sample aggregation, (ii) Łukasiewicz t-norm forward composition and residuum-based backward, (iii)  pullback verification for hard constraints, and (iv) explicit $\alpha$-cut plan acceptance.

\subsection{Metrics and Evaluation Protocol}
\label{sec:metrics}
\paragraph{Success Rate \%.}
We report the fraction of instances where the produced plan/output succeed in  executability checks.
\begin{itemize}
    \item \textbf{PDDL3}: plan valid under domain dynamics or violates hard constraints.
    \item \textbf{RecipeNLG-Subs}: follow all constraints (e.g., allergen forbidden ingredient appears; required equipment absent; time/resource constraints violated).
\end{itemize}


\section{Results}
\label{sec:results}

\subsection{Main results}

\begin{table}[t]
\centering
\small
\begin{tabular}{l|c|cc}
\toprule
& \multicolumn{1}{c|}{\textbf{PDDL3 }} & \multicolumn{2}{c}{\textbf{RecipeNLG-Subs}} \\
\textbf{Method} & \textbf{Success} & \textbf{BLEU} & \textbf{Success}   \\
\midrule
SGPlan & 90.6  & -- & -- \\
LPPG & 21.3 (P only)  & -- & -- \\
MIPS-XXL & 25.1  & -- & -- \\
\midrule
SubDB-Retrieve & -- & 92.7 & 67.8 \\
ReAct & --  & 90.8 & 52.1 \\
LLM & -- & 90.4 & 20.4 \\
\midrule
\textbf{FCP} & 85.2  & 90.5 & 83.6 \\
\bottomrule
\end{tabular}
\caption{Main results. Violation measures hard-constraint failures; BLEU measures plan realization similarity to reference.}
\label{tab:main}
\end{table}

On the \textbf{PDDL3} suite, the classical planner \textbf{SGPlan} achieves the highest success rate (90.6), which is consistent with its purpose-built optimization for crisp dynamics with preferences/utility objectives. \textbf{FCP} attains a competitive success rate (85.2) but remains below SGPlan, suggesting that our fuzzy-composition machinery is not intended to replace specialized symbolic solvers on fully formal PDDL3 domains, and may incur overhead or mild modeling mismatch when preferences are already explicitly encoded. \textbf{LPPG} and \textbf{MIPS-XXL} perform substantially worse in our configuration (21.3 and 25.1), and we therefore treat them as additional reference points rather than the primary ``best classical'' comparator.

On \textbf{RecipeNLG-Subs}, we observe a clear separation between \emph{textual similarity} and \emph{constraint-valid planning success}. Retrieval-driven substitution (\textbf{SubDB-Retrieve}) yields the highest BLEU (92.7), which is expected because it minimally edits recipes toward reference realizations; however, its success rate (67.8) indicates that high surface overlap does not reliably imply validity under the constraint checker. Pure prompting (\textbf{LLM}) and \textbf{ReAct} also obtain relatively high BLEU (90.4--90.8) but much lower success (20.4 and 52.1), suggesting they often produce fluent, reference-like text while violating hard constraints (e.g., forbidden ingredients, inconsistent substitutions, or missing required steps/equipment). In contrast, \textbf{FCP} achieves the highest success rate (83.6) while producing lower BLEU (90.5), consistent with the hypothesis that enforcing crisp validity while composing graded satisfactions can lead to more substantive edits than retrieval-only baselines, even when the output remains reasonably close to reference realizations.

\subsection{Ablation analysis}
Table~\ref{tab:ablation} suggests that FCP’s performance is driven by the combination of stable membership grounding and mechanisms that mitigate \L{}ukasiewicz degradation. First, removing aggregation ($k{=}1$) consistently reduces Success on both benchmarks, while only slightly affecting BLEU. This pattern indicates that aggregation mainly improves \emph{decision reliability} (accept/reject and action selection under graded applicability) rather than surface realization quality. Second, disabling chunking causes the largest drop with only a modest BLEU decrease, consistent with chunking acting as the primary mitigation for multi-step degradation: without reducing effective plan depth, many otherwise reasonable plans fail the quality threshold after composition. Finally, replacing \L{}ukasiewicz with the G\"odel (min) t-norm reduces Success while leaving BLEU essentially unchanged. This suggests that min-composition is less aligned with the target notion of \emph{accumulated} quality loss: it preserves only the worst step and does not provide the same controllable trade-off between partial satisfaction and plan length, which in turn harms success rate even when the generated text remains fluent.

\begin{table}[t]
\centering
\small
\begin{tabular}{l|c|cc}
\toprule
& \multicolumn{1}{c|}{\textbf{PDDL3}} & \multicolumn{2}{c}{\textbf{RecipeNLG-Subs}} \\
\textbf{Variant} & \textbf{Success}  & \textbf{BLEU} & \textbf{Success} \\
\midrule
FCP (full) & 85.2  & 90.5 & 83.6 \\
\quad -- $k{=}1$ (no aggregation) & 74.5  & 89.1 & 78.5 \\
\quad -- chunking & 48.3  & 88.6 & 51.0 \\
\midrule
\quad t-norm: Gödel (min) & 69.7 & 90.6 & 72.8 \\
\bottomrule
\end{tabular}
\caption{Ablations Result}
\label{tab:ablation}
\end{table}

\subsection{Degradation analysis}
\begin{figure}[t]
    \centering
    \includegraphics[width=0.7\linewidth]{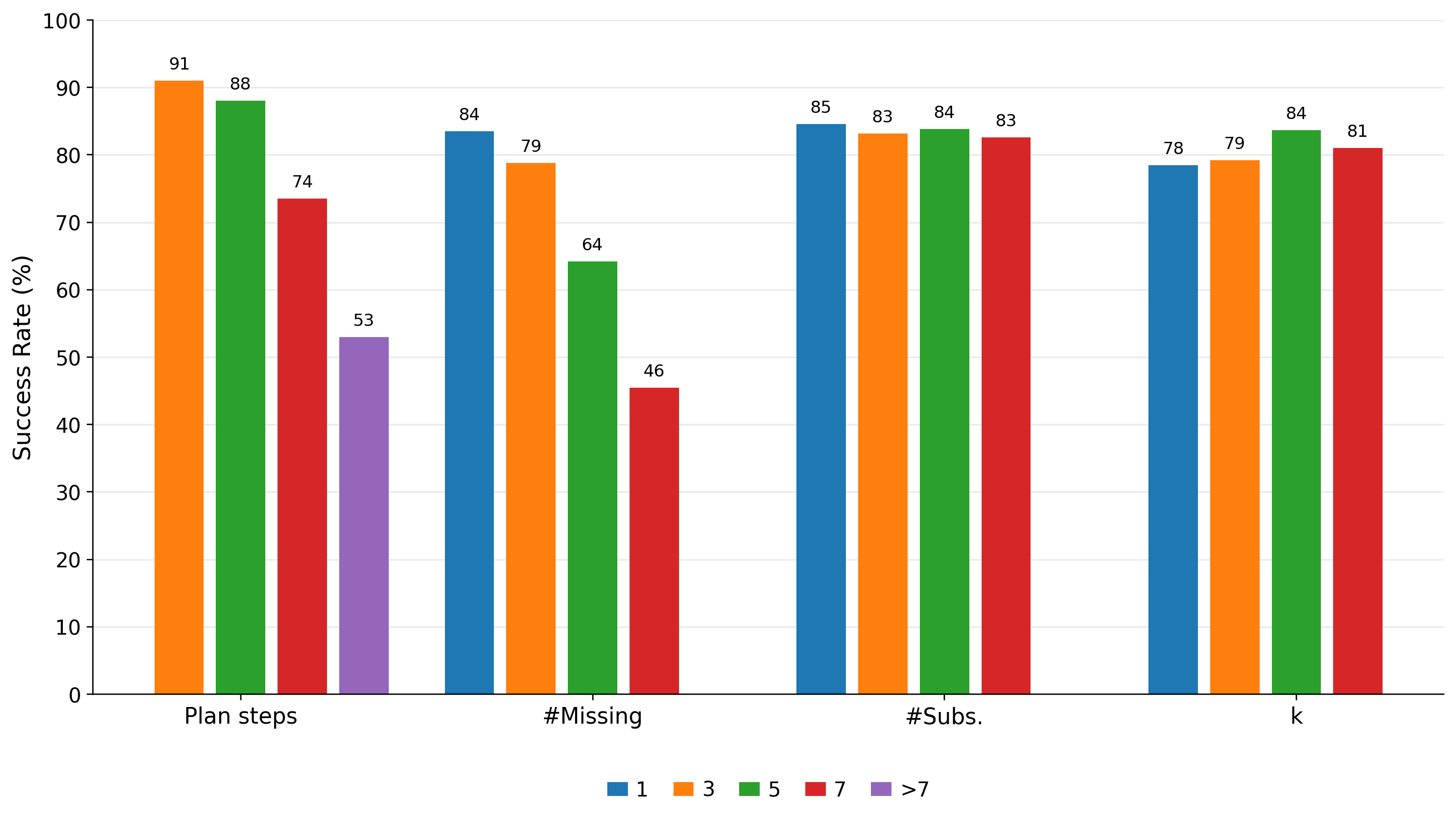}
    \caption{\textbf{Degradation drivers in FCP.} Success rate (\%) as a function of plan length, number of missing requirements, number of substitute candidates, and grounding aggregation size $k$ (color denotes the x-axis bin: 1/3/5/7/$>7$ or 1/3/5/7 depending on the factor). Performance degrades sharply with longer plans and more missing items, while varying the number of candidates has minimal impact; increasing $k$ improves robustness up to $k{=}5$ with diminishing returns thereafter.}

    \label{fig:miss_precondition}
\end{figure}

We analyze how FCP success varies with four drivers of cumulative difficulty: plan length, the number of missing requirements, the number of substitute candidates, and the grounding aggregation size $k$. The result is shown in Figure~\ref{fig:miss_precondition}. Success decreases markedly with longer plans (especially beyond 7 steps) and with more missing items, consistent with compounding degradation under t-norm composition and additional repair decisions. In contrast, varying the number of candidate substitutes has a comparatively small effect, suggesting that ranking among alternatives is not the primary bottleneck once a reasonable candidate set is available. Finally, aggregation improves robustness from $k{=}1$ to $k{=}5$, with diminishing returns at larger $k$, supporting $k{=}5$ as a practical default.

\subsection{Model Scale and Architecture}
FCP achieves consistently high success across backbones, with only moderate variation (approximately 80--92\%).Stronger backbones yield small but clear gains, with the best performance on Claude~3.5 and competitive results from 14B-class models (DeepSeek-R1-14B, Qwen3-14B).

\begin{figure}[t]
    \centering
    \includegraphics[width=0.6\linewidth]{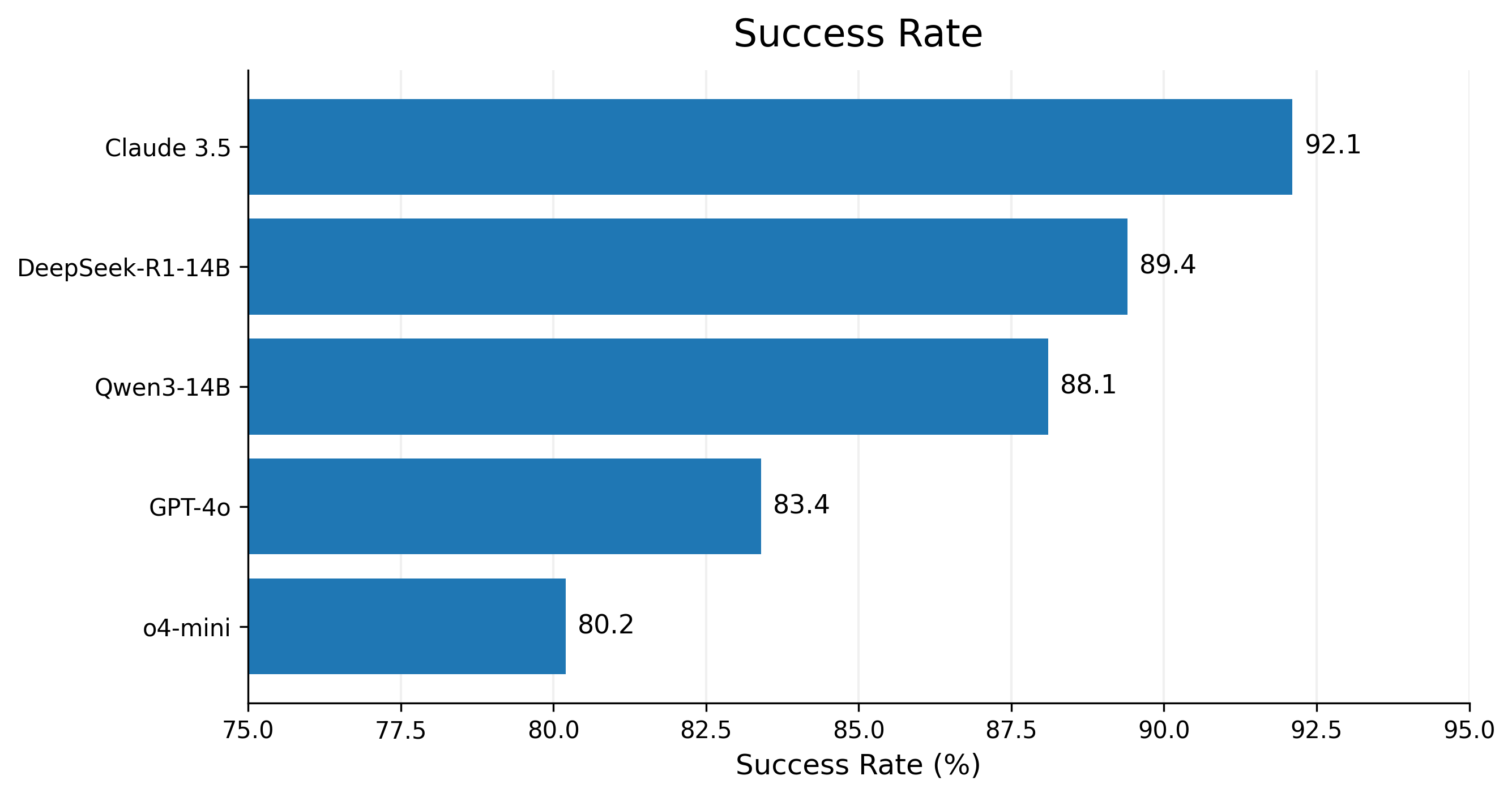}
    \caption{\textbf{Influence of backbone.} FCP maintains high success across backbones (80--92\%), with modest gains from stronger models.}
    \label{fig:2}
\end{figure}

\section{Conclusion}

We introduced Fuzzy Category-theoretic Planning (FCP), which augments categorical planning with graded morphism applicability and \L{}ukasiewicz composition to reason explicitly about semantic vagueness while preserving hard executability through pullback-style verification. Across PDDL3 preference/oversubscription planning and RecipeNLG-Subs missing-substitute adaptation, FCP improves success in settings where partial satisfaction must be composed and controlled. Ablations show that aggregation, chunking, and the choice of t-norm are important for robustness, and our degradation analysis clarifies when nilpotent composition limits effective plan depth. 

\section{Limitations}
First, automated dataset construction can introduce artifacts (e.g., systematic substitute distributions) that inflate BLEU or simplify constraint checking. We mitigate this by stratifying by ambiguity and constraint density, and by reporting performance across bins. Second, BLEU can under-reward semantically valid paraphrases; future work should add embedding-based or semantic similarity metrics and user studies. Third, preference/oversubscription objectives differ from semantic vagueness in natural language; we view PDDL3 as a conservative proxy that still captures the key desideratum: non-binary optimization under crisp executability.

\bibliography{custom}

\appendix

\end{document}